\documentclass[final]{cvpr}

\usepackage{times}
\usepackage{epsfig}
\usepackage{graphicx}
\usepackage{breqn}
\usepackage{amsmath}
\usepackage{amssymb}

\usepackage{booktabs}       

\usepackage{algorithm}
\usepackage{algpseudocode}

\usepackage[pagebackref=true,breaklinks=true,colorlinks,bookmarks=false]{hyperref}

\def\cvprPaperID{3609} 
\def\confYear{CVPR 2021}

\begin{document}

\title{Skeleton Merger: an Unsupervised Aligned Keypoint Detector}

\author{Ruoxi Shi, Zhengrong Xue, Yang You, Cewu Lu\thanks{Cewu Lu is corresponding author. He is also the member of Qing Yuan Research Institute and MoE Key Lab of Artificial Intelligence, AI Institute, Shanghai Jiao Tong University, China}\\
Shanghai Jiao Tong University\\
{\tt\small eliphat,  xuezhengrong,  qq456cvb,  lucewu@sjtu.edu.cn}
}

\maketitle

\begin{abstract}
   Detecting aligned 3D keypoints is essential under many scenarios such as object tracking, shape retrieval and robotics. However, it is generally hard to prepare a high-quality dataset for all types of objects due to the ambiguity of keypoint itself. Meanwhile, current unsupervised detectors are unable to generate aligned keypoints with good coverage. In this paper, we propose an unsupervised aligned keypoint detector, Skeleton Merger, which utilizes skeletons to reconstruct objects. It is based on an Autoencoder architecture. The encoder proposes keypoints and predicts activation strengths of edges between keypoints. The decoder performs uniform sampling on the skeleton and refines it into small point clouds with pointwise offsets. Then the activation strengths are applied and the sub-clouds are merged. Composite Chamfer Distance (CCD) is proposed as a distance between the input point cloud and the reconstruction composed of sub-clouds masked by activation strengths. We demonstrate that Skeleton Merger is capable of detecting semantically-rich salient keypoints with good alignment, and shows comparable performance to supervised methods on the KeypointNet dataset. It is also shown that the detector is robust to noise and subsampling. Our code is available at \href{https://github.com/eliphatfs/SkeletonMerger}{https://github.com/eliphatfs/SkeletonMerger}.
\end{abstract}

\section{Introduction}

\begin{figure}[h]
  \centering
  \includegraphics[width=0.8\linewidth]{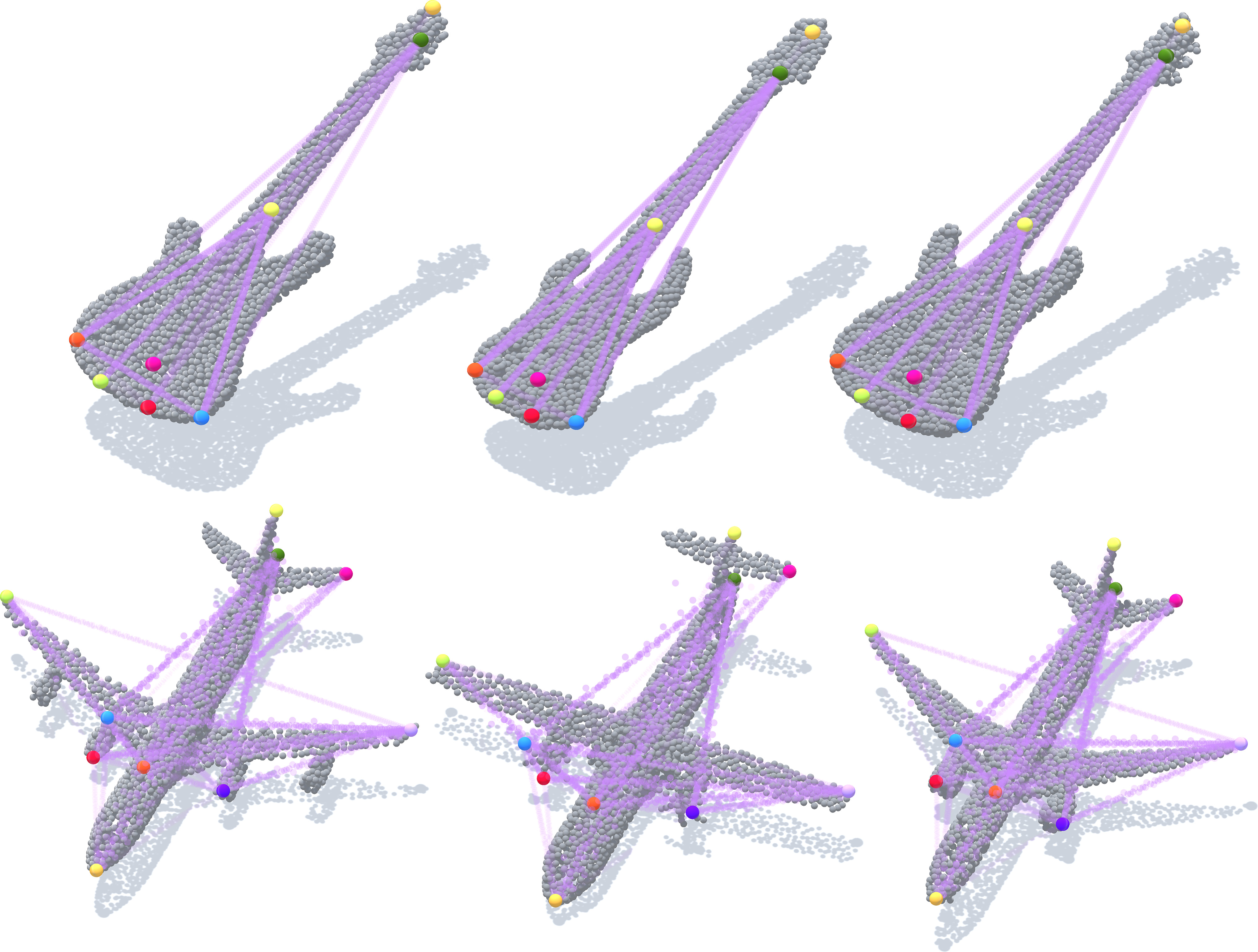}
  \caption{\textbf{Examples of detected aligned keypoints and the corresponding skeletons from our detector.} Skeleton is shown in purple lines. Objects are from ModelNet40 \cite{wu20153d} and ShapeNetCoreV2 \cite{shapenet2015} datasets.}
  \label{intropic}
\end{figure}

Being able to fully understand an object is arguably the ultimate goal of computer vision. For 3D point clouds, detecting semantic keypoints is currently the most promising and widely adopted approach. \cite{weinmann2016reconstruction,tamas20143d} Keypoints are crucial to the success of many vision applications such as object tracking, shape registration and in robotics \cite{mian2006three,wang2018learning,besl1992method,choi2010real}. In many actual cases where objects from the same category are compared, we desire keypoints to be not only accurately located but also aligned within a certain category for performing high-level vision tasks such as 3D object recognition and reconstruction. \cite{yang2017performance}

However, to decide whether a point contains semantics requires high-level intelligence, as `semantics' itself is ambiguously defined. Different people would present different understandings of semantic points. Therefore, very limited human annotated quality data are available so far \cite{you2020keypointnet}. Consequently, supervised methods can only deal with very limited range of objects covered in the datasets despite their success on many other tasks.

Most unsupervised methods, either traditional hand-crafted ones \cite{sipiran2011harris,Zhong2009IntrinsicSS} or deep learning-based \cite{li2019usip}, take advantage of geometric properties to detect keypoints. While stable, these keypoints are often not rich in semantics, and the coverage of keypoints on the input point cloud is generally low, especially under a small number of points, which limits their performance in downstream tasks. Moreover, the keypoints detected are neither ordered nor aligned. A very recent approach \cite{fernandez2020unsupervised} can learn aligned 3D keypoints by decomposing keypoint coordinates into a low-rank non-rigid shape representation. However, in categories like the airplane where objects do not necessarily share highly similar geometric shapes, its performance dramatically declines.

In view of the challenges above, we propose Skeleton Merger, an unsupervised keypoint detector that can extract salient and aligned keypoints. As its name suggests, Skeleton Merger attempts to reconstruct a point cloud from its skeleton. Some examples of detected keypoints and the skeletons are shown in Fig.\ \ref{intropic}.

\begin{figure}[t]
  \centering
  \includegraphics[width=\linewidth]{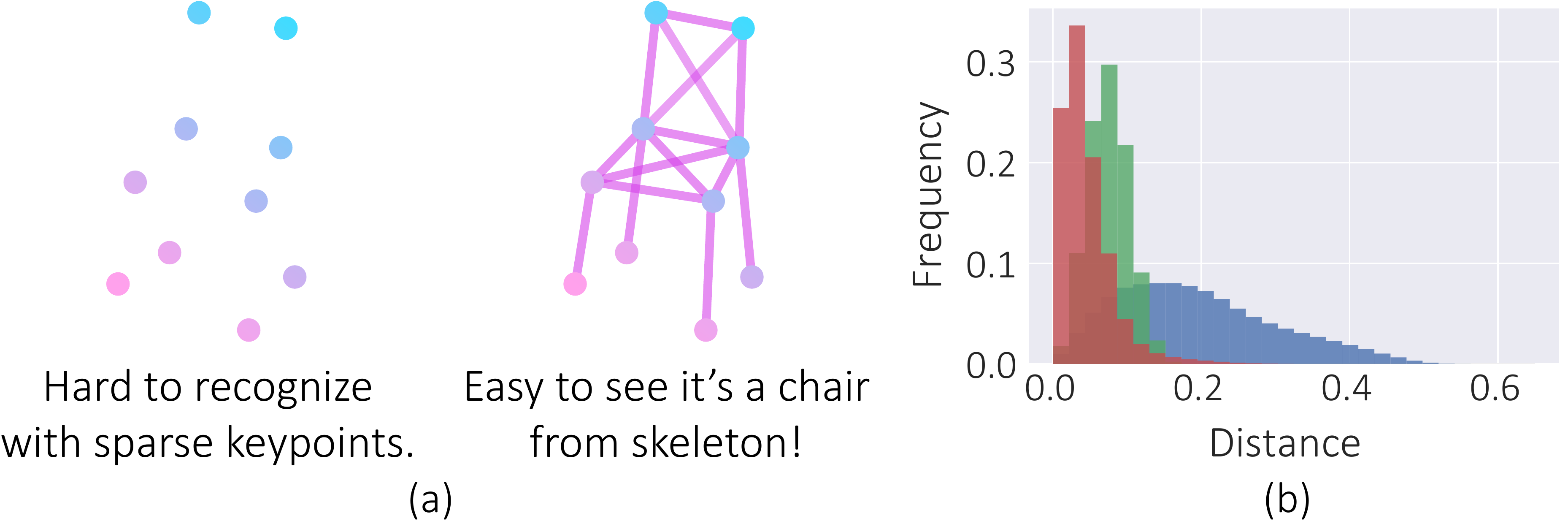}
  \caption{\textbf{Describe objects with skeletons.} (a) We observe that human vision can easily distinguish objects with skeletons, but it is harder with keypoints only. (b) Histogram of nearest neighbour distances of points in point clouds to its skeleton (red), large quantity of (3200) uniformly sampled points (green) in the bounding box and keypoints (blue).}
  \label{plane_kp_sketch}
\end{figure}

Utilizing skeleton to represent a point cloud is inspired by the skeleton extraction problem \cite{cornea2007curve} in traditional graphics. We hold the belief that skeleton is a better presentation method than that adopted by current deep learning-based frameworks both intuitively and statistically. Qualitatively speaking, we observe that human vision tends to use a `joint-skeleton' cognitive pattern to recognize things, as is shown in Fig.\ \ref{plane_kp_sketch} (a). While accurate and descriptive semantics is mostly provided by keypoints, with some discrete keypoints as joints alone, it is merely impossible for human beings to recognize objects. However, if we introduce some auxiliary line segments connecting certain keypoints together to form a skeleton, humans can easily distinguish the object. Quantitatively speaking, statistical results in Fig.\ \ref{plane_kp_sketch} (b) show that points of a point cloud are in general closer to those of a skeleton (red), compared with keypoints (blue) or uniform sampling inside the bounding box (green). This indicates skeleton can better fit with local shape features of the original point cloud.

On implementation level, our Skeleton Merger follows a deep Autoencoder architecture. The encoder generates keypoint proposals. It also predicts activation strengths of edges between keypoints. The decoder generates a skeleton from these results, which is essentially a graph of keypoints. It performs uniform sampling on each edge, adds distinct activation strength and offset to refine the shape. In this way, each skeleton edge is essentially a small sub-cloud. The decoder finally merges them altogether to form a reconstruction point cloud. Noticing that the order of skeleton edges is predefined by the encoder, the alignment of keypoints is therefore considerably improved.

The crucial problem now becomes how to construct a loss function that can evaluate how well the refined skeleton reconstructs the original point cloud. Following the idea of traditional Chamfer loss \cite{barrow1977parametric}, which is the sum of forward and backward losses, we come up with Composite Chamfer Distance (CCD), which is the sum of fidelity (forward) and coverage (backward) losses. CCD measures the distance between the input and the reconstructed point cloud composed of many sub-clouds masked by activation strengths.

Experimental results show that Skeleton Merger is capable of detecting semantically-rich salient keypoints with good alignment. Our detector achieves significantly better performance than current unsupervised detectors on the KeypointNet \cite{you2020keypointnet} dataset. In fact, its performance is even comparable to supervised methods. Results also show that our detector is robust to noise and subsampling.


\section{Related work}
\paragraph{Curve skeleton extraction}
In the context of computer graphics, skeleton refers to the 1D structure which is a simplified representation of the geometry of a 3D object. Cornea \etal \cite{cornea2007curve} provided an overview of curve skeleton algorithms \cite{au2008skeleton,ma2003skeleton,shen2016object} and their applications. In our paper, however, skeleton refers to a graph of keypoints that represents topology of the object. The purpose of our skeleton is not only to provide a rough geometric shape of the original object, but to help improve alignment of keypoints.

\paragraph{Deep learning on 3D point clouds}
Currently, various deep learning-based techniques that consume point clouds \cite{liu2019relation,qi2017pointnet,qi2017pointnetplusplus,xu2018spidercnn,wu2019pointconv,kanezaki2018rotationnet} have been developed. They initially aim at basic tasks such as classification and segmentation, but can be adapted for more high-level 3D perception tasks like point cloud registration, pose estimation and 3D reconstruction \cite{guo2020deep,elbaz20173d,zeng2017multi,han2019image}.
PointNet \cite{qi2017pointnet} proposed by Qi \etal is a pioneering work that first enables neural networks to directly process raw point cloud data. In PointNet, the input points pass through per-point multi-layer perceptron and a symmetric max pooling operation, ending up with a global feature. The global feature is then used for various downstream tasks.
However, PointNet only takes notice of the global information, neglecting local details. Thus, Qi \etal extended PointNet to PointNet++ \cite{qi2017pointnetplusplus}, where PointNet is applied hierarchically on different spatial scales for better performance on point clouds. Our encoder utilizes PointNet++ as a point cloud processing block.

\paragraph{Unsupervised 3D keypoint detectors}
Currently, most 3D keypoint detectors remain to be hand-crafted. Popular hand-crafted detectors such as Harris 3D \cite{sipiran2011harris}, ISS \cite{Zhong2009IntrinsicSS}, HKS \cite{sun2009concise}, SHOT \cite{tombari2010unique} take advantage of geometric properties to select most salient keypoints. As geometric characteristics are quite complex in 3D objects, most keypoints detected are neither semantically salient nor well aligned.

\begin{figure*}[t]
  \centering
  \includegraphics[width=0.8\linewidth]{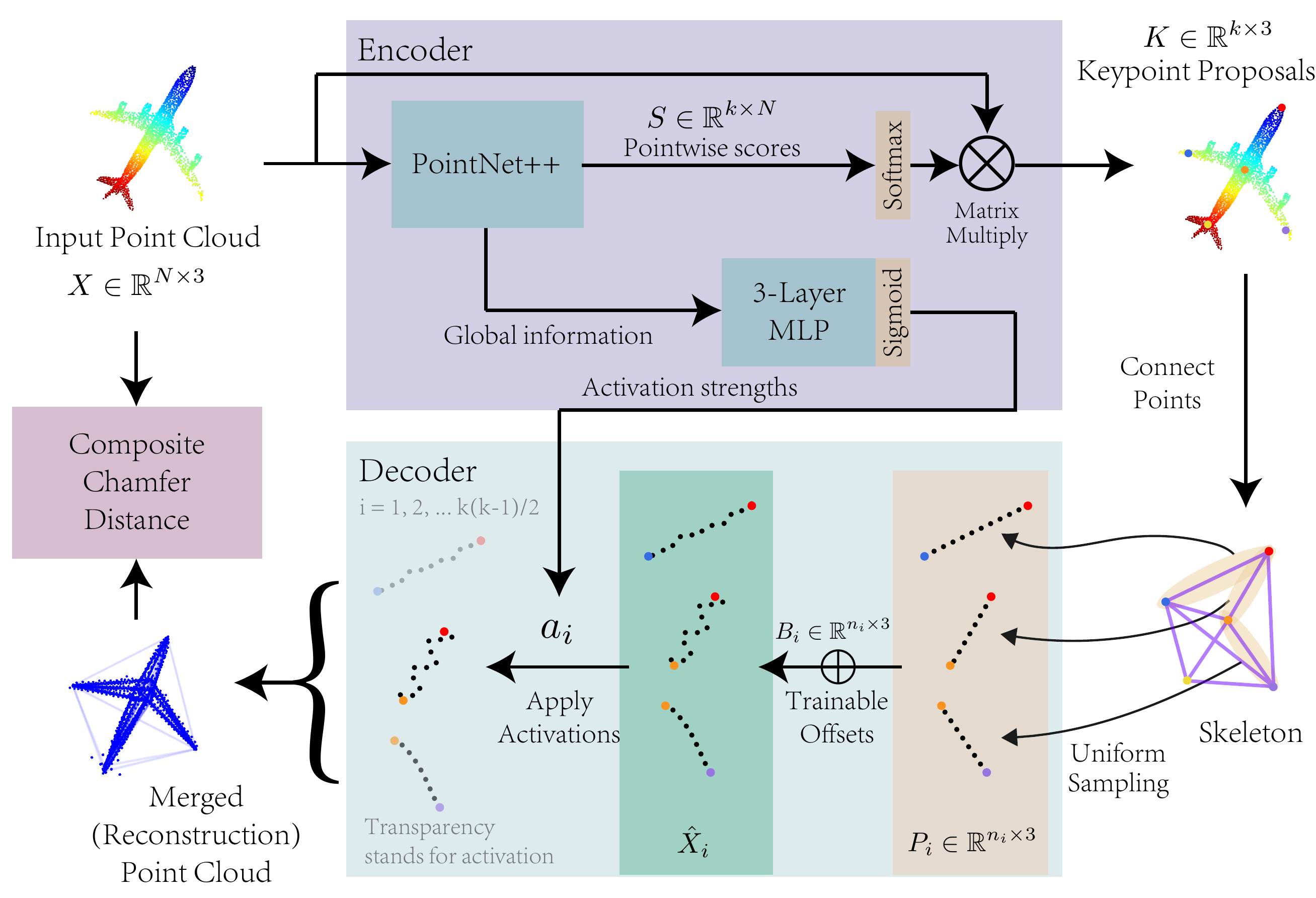}
  \caption{\textbf{Pipeline of Skeleton Merger.} The encoder takes $n$ points as input and utilizes a PointNet++ network to generate keypoint proposals and activation strengths. `MLP' stands for multi-layer perceptron. Batch-norm and ReLU are used for the MLP. The decoder samples on edges of the skeleton, refines them by adding positional offsets that are directly optimized as parameters, and merges the refined skeleton edges with the activation strengths. Composite Chamfer Distance is applied between the reconstruction result and the input point cloud.}
  \label{pipeline}
\end{figure*}

To the best of our knowledge, USIP \cite{li2019usip} is the first learning-based 3D keypoint detector. USIP takes advantage of probabilistic chamfer loss that greatly enhances repeatability. However, while keypoints detected are stable under geometric transformations, they may have poor coverage on the point cloud, especially when the number of keypoints is small. Furthermore, keypoints detected are unordered and not aligned.
Very recently, Fernandez \etal \cite{fernandez2020unsupervised} propose an unsupervised approach that can learn aligned 3D keypoints through decomposing keypoint coordinates into a low-rank non-rigid shape representation. This leads to degraded performance on classes where objects are not necessarily similar in geometry such as guitars and airplanes of irregular shapes. Moreover, all shapes are assumed to be axisymmetric in their method, while our detector needs no symmetry prior, so it can be applied to a wider range of objects.

\section{Methods}

Now we propose the unsupervised aligned keypoint detector, Skeleton Merger, which is based on a deep Autoencoder framework. The full pipeline is shown in Fig.\ \ref{pipeline}.

The encoder first proposes an ordered list of $k$ keypoints $K \in \mathbb{R}^{k \times 3}$. Connecting each pair of keypoints, a rough skeleton of the input point cloud is generated, composed of $\mathrm{C}_k^2 = k(k-1)/2$ edges. The skeleton is further refined into $k(k-1)/2$ point clouds through uniform sampling and addition with a sequence of learned offset vectors. These clouds are masked by $k(k-1)/2$ activation strengths, which also come from the encoder, and merged into a final reconstruction point cloud at the output of the decoder. The Composite Chamfer Distance (CCD) is applied between the merged reconstruction point cloud and input point cloud to guide the unsupervised learning process.

In the following subsections, we will introduce in detail each module in the pipeline.

\subsection{Skeleton and activation strengths}
\label{sacts}
The key of the Skeleton Merger architecture lies in the utilization of the skeleton to reconstruct objects.

Directly reconstructing the original point cloud from keypoints is essentially transforming a point set into a larger one, which by its nature is unordered. Furthermore, keypoints only contain sparse semantic information, and experiments show that reconstruction by this method may lead to poor coverage of the original point cloud. In contrast, the skeleton provides a basic geometric shape of the original object, instead of sparse semantics. Meantime, it can be easily constructed with alignment by connecting pairs of keypoints in order.

There are edges in the skeleton, however, that do not actually exist in the original point cloud. They will bring unwanted noise to the reconstruction. A mechanism is thus required for masking out these non-existing edges. So together with the skeleton, a list of activation strength values are predicted by the model. These values indicate existence of edges in the skeleton. This step requires the model to learn the skeleton topology of the reconstruction target object, which, combined with ordered construction of the skeleton edges, improves alignment of keypoints.

\subsection{Encoder, the keypoint proposer}
In the encoder, a PointNet++ \cite{qi2017pointnetplusplus} network is first applied to obtain $k$ pointwise scores $S \in \mathbb{R}^{k \times N}$ of the input point cloud. The pointwise scores are then activated by a softmax function. A weighted average of the input point cloud $X \in \mathbb{R}^{N \times 3}$ is computed from $\verb+softmax+(S)$ and $X$ to form the final keypoint proposals $K \in \mathbb{R}^{k \times 3}$. The weighted average is implemented by a matrix multiplication, shown in Fig.\ \ref{pipeline}.

Besides $k$ keypoints, the existence of $k(k-1)/2$ edges between each pair of keypoints is encoded as well. A 3-layer MLP (multi-layer perceptron) accepts the \emph{global feature vector} generated by the PointNet++ network and predicts $k(k-1)/2$ sigmoid-activated activation strengths $a \in \mathbb{R}^{k(k-1)/2}$. The activation strengths are used to mask skeleton edges before the merging stage in decoder.

\subsection{Decoder, the skeleton refiner}
The decoder takes as input the keypoints $K \in \mathbb{R}^{k \times 3}$ proposed by the encoder. A uniform sampling operation is performed on the edges to get $k(k-1)/2$ small point clouds, $P_1, P_2, \dots, P_{k(k-1)/2}$. The number of points sampled $n_i$ is in proportion to the length of each edge.

For each small point cloud $P_i \in \mathbb{R}^{n_i\times3}$, pointwise position offset $B_i \in \mathbb{R}^{n_i\times3}$ is added to the initial points sampled as a refinement to the skeleton formed by straight lines. $B_i$'s are directly optimized as parameters of the network. In order to keep the refinement localized, a Ridge (L2) regularization is imposed on the learnt position offsets.

The $k(k-1)/2$ refined point clouds $\hat{X}_i = P_i + B_i$ are merged into a single point cloud with the activation strengths $a_i$ from the encoder. The CCD is then applied between the reconstruction point cloud and the input point cloud $X$ as the loss to guide the training.

\subsection{Composite Chamfer Distance}
To put the reconstruction process into practice, it is essential to establish a loss function between the input point cloud ${X}$ and the reconstruction point cloud composed of different parts $\hat{X}_i$ masked by activation strengths $a_i$. We generalize the Chamfer Distance to take into account the activation strengths, and proposes the Composite Chamfer Distance (CCD).

Similar to the regular Chamfer Distance, CCD is a sum of fidelity (forward) and coverage (backward) losses. However, the reconstruction result is composed of several sub-clouds, while the input is simply one large point cloud. This asymmetry leads to asymmetry in designs of fidelity and coverage losses, making it to fit the nature of the problem.

The fidelity loss is a straightforward extension to the Chamfer Distance where the activation strength is applied to each sub-cloud, as shown in Eq.\ \ref{ccd_forward}:
\begin{equation}
\label{ccd_forward}
\mathcal{L}_f = \sum_i a_i \sum_{\hat{p} \in \hat{X}_i} \min_{p_0 \in X} \left|\left|\hat{p} - p_0\right|\right|_2.
\end{equation}

Then it comes to the coverage loss. If we do the same simple extension to the Chamfer Distance, activation strengths $a_i$ will go to zero when the loss gets minimized, which prevents the model from learning anything meaningful. The problem is that, a reconstruction point with a small $a_i$ value does not contribute to coverage as much as one with a large $a_i$ value. Therefore, more points than only the one with minimal distance should be considered if its activation strength is not large enough.

\begin{figure}
  \includegraphics[width=0.98\linewidth]{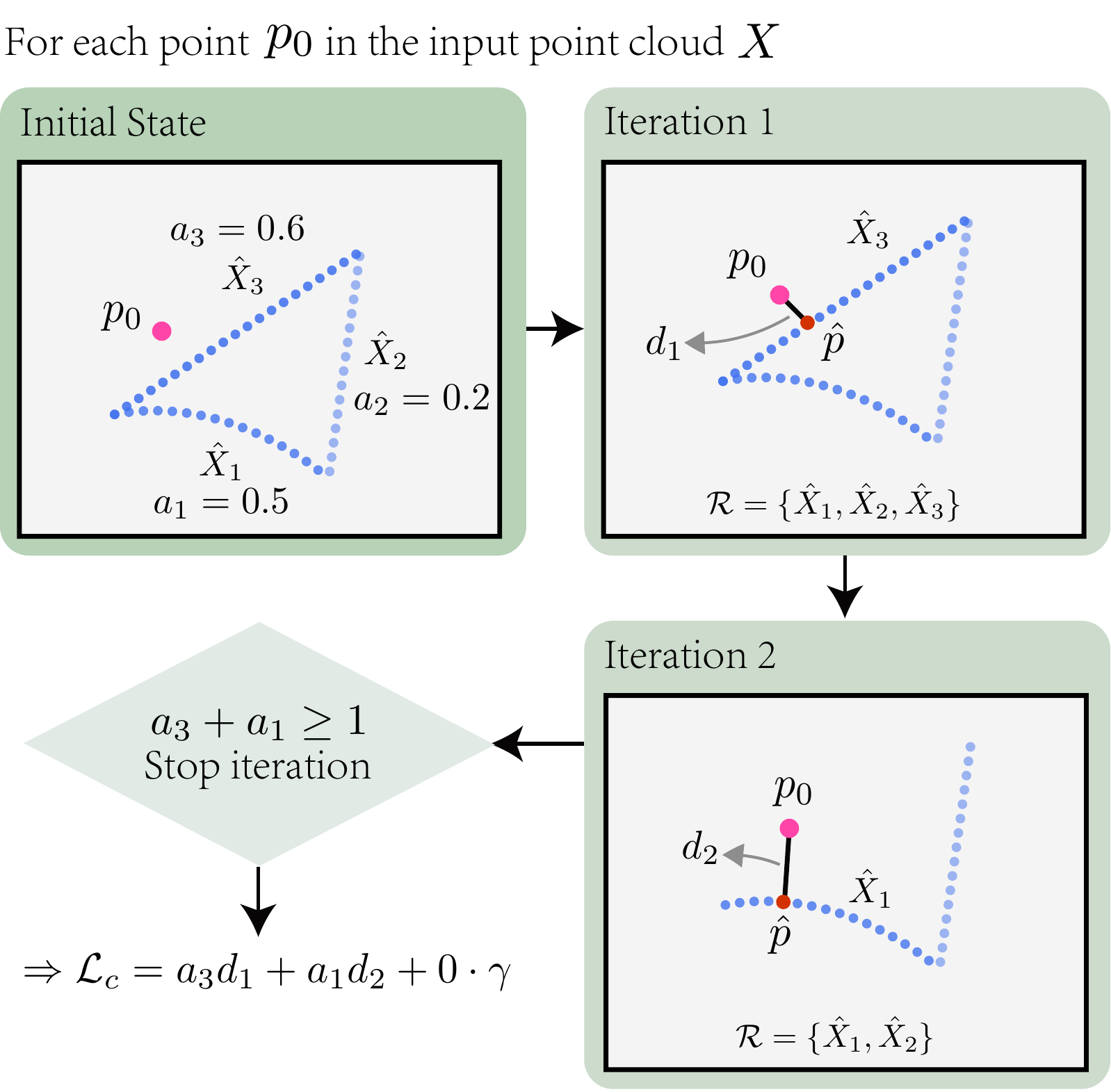}
  \centering
  \caption{\textbf{An illustrating example of the coverage loss algorithm.} The input cloud is a curved poly-line in the example. Currently the focus $p_0$ is the marked pink point. In each iteration, the nearest point of $p_0$ in $\mathcal{R}$ is selected and the distance is computed. Then the small point cloud that contains the selected point is discarded from $\mathcal{R}$ and the next iteration starts. The iteration stops when selected $a_i$'s sum up to 1 (as shown), or when $\mathcal{R}$ is empty, in which case $\gamma$ would be applied as a penalty for the rest of $w$. See the text for more details about coverage loss.}
  \label{ccdlcfig}
\end{figure}

In view of this, we come up with the following coverage loss, which involves point-wise sorting of sub-clouds and weighted averaging instead of a simple minimum. The algorithm to generate the coverage loss is shown in Alg.\ \ref{ccd_backward}. An illustrating example is shown in Fig.\ \ref{ccdlcfig}.

\begin{algorithm}[h]
\caption{Coverage loss of CCD}
\label{ccd_backward}
\hspace*{\algorithmicindent} \textbf{Input}: $X$, $\hat{X}_1 \dots \hat{X}_{k(k-1)/2}$, $a_1 \dots a_{k(k-1)/2}$ \\
\hspace*{\algorithmicindent} \textbf{Parameter}: $\gamma$ \\
\hspace*{\algorithmicindent} \textbf{Output}: $\mathcal{L}_c$
\begin{algorithmic}[1]
\For {$p_0 \in X$}
  \State $\mathcal{R} \leftarrow \bigcup\ \{\hat{X}_1 \dots \hat{X}_{k(k-1)/2}\}$
  \State $w \leftarrow 0$
  \While {$w < 1$ and $\mathcal{R} \ne \emptyset$}
    \State Find $\hat{p}$ by $\min_{\hat{p} \in \mathcal{R}} \left|\left|\hat{p} - p_0\right|\right|_2$
    \State Find $i$ with $\hat{p} \in \hat{X}_i$
    \State $\mathcal{L}_c \leftarrow \mathcal{L}_c + a_i \left|\left|\hat{p} - p_0\right|\right|_2$
    \State $w \leftarrow w + a_i, \mathcal{R} \leftarrow \mathcal{R} \setminus \hat{X}_i$
  \EndWhile
  \If {$w < 1$} \State $\mathcal{L}_c \leftarrow \mathcal{L}_c + \gamma (1 - w)$ \EndIf
\EndFor
\end{algorithmic}
\end{algorithm}

Each point in the input point cloud is treated separately (the outer `for' loop starting at line 1) with an iterative process: the point with minimal distance in the collection of sub-clouds is selected (line 5 and 6), and the distance multiplied with the activation strength is the contribution of the current iteration to the coverage loss (line 7), after which the entire sub-cloud is removed from the collection under consideration (line 8). The iteration stops after activation strengths sum up to $1$ (notice the $w < 1$ condition in the while loop starting at line 4). If it never reaches $1$, a high penalty of $\gamma$ is imposed (line 10 to 12). $\gamma$ is set to $20.0$ in the experiments.

Intuitively, sub-clouds that contribute more to the coverage have small backward Chamfer Distance, and due to the cutting mechanism, the activation strengths of these parts will get larger. The fidelity loss, in contrast, reduces the activation strengths of non-existing edges in the skeleton. By applying CCD as loss, the network is forced to generate a set of sub-clouds with reasonable activation values, which enables the training process.

Meantime, alignment is assured since the activation-generating MLP only sees global information about the point cloud. The MLP learns essentially the topology of target model skeletons, and a wrong ordering of keypoints will lead to high fidelity and coverage losses.

The final CCD loss is a weighted sum of the two parts, fidelity loss $\mathcal{L}_f$ and coverage loss $\mathcal{L}_c$, as shown in Eq.\ \ref{ccdf}. The weight coefficients $\lambda_f$ and $\lambda_c$ can be tuned. They default to be the same in the experiments.
\begin{equation}
\mathcal{L} = \lambda_f \mathcal{L}_f + \lambda_c \mathcal{L}_c.
\label{ccdf}
\end{equation}

\section{Results}
In this section, we evaluate keypoint saliency, alignment and repeatability of the proposed Skeleton Merger, and give some qualitative results of the keypoint detector.

\subsection{Metric for alignment with human annotations}
Currently, there lacks a metric for evaluating correspondence between two sets of semantic labels for keypoints, where a consistent label is given to keypoints with the same semantics in each set, but the relation between labels of two sets is unknown. We propose Dual Alignment Score (DAS) metric for this evaluation.

\begin{figure}[h]
  \centering
  \includegraphics[width=0.95\linewidth]{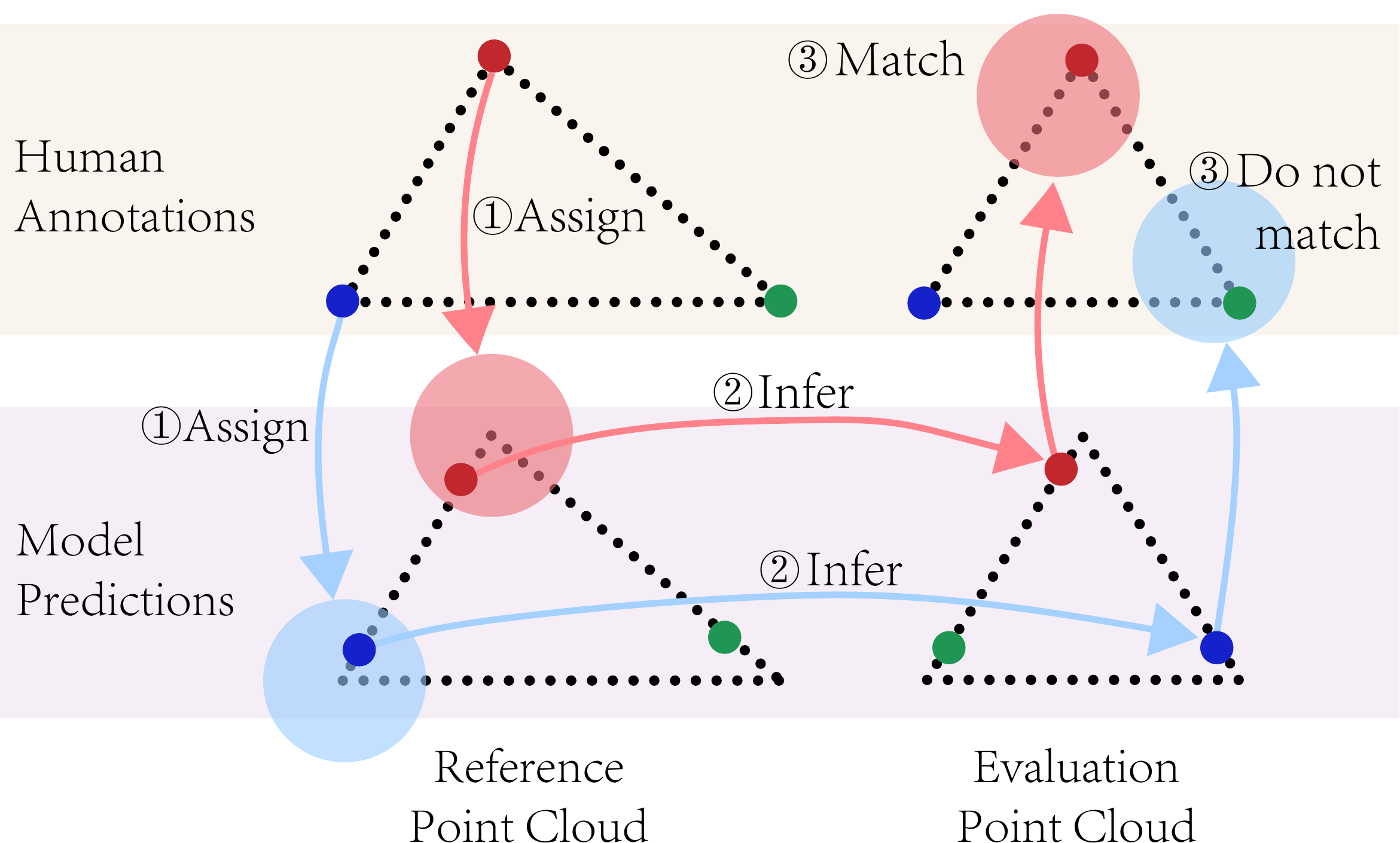}
  \caption{\textbf{An illustration of DAS computation process.} Only one matching direction from predicted semantic labels to annotations is shown. We first assign corresponding labels by finding the nearest neighbour of predicted keypoints on annotations of the reference point cloud. Then the human annotated keypoints with same semantic labels are inferred for the evaluation point cloud. Finally, we see if the nearest neighbour of the inferred annotations are of same predicted labels, and compute an accuracy score.}
  \label{dasi}
\end{figure}

\paragraph{Dual Alignment Score}
In order to evaluate whether our keypoints are consistent in semantics, we propose Dual Alignment Score (DAS). In DAS calculation,  a reference point cloud is used for semantic assignment and another model is used for the actual evaluation. On the reference model, we assign each predicted keypoint with the semantic index of the closest point from human annotations. Then on the other point cloud, the corresponding predicted point is found since our keypoints are aligned. A score is calculated by the mean accuracy whether the closest human annotated keypoint of this point is of the same semantic index. In the opposite direction, order of our predicted keypoints is used to assign semantic labels for human annotations, and the process is repeated. By averaging scores in these two directions we get the Dual Alignment Score (DAS). Fig.\ \ref{dasi} shows an illustration of the DAS computation.

\begin{figure*}[t]
  \centering
  \includegraphics[width=0.8\linewidth]{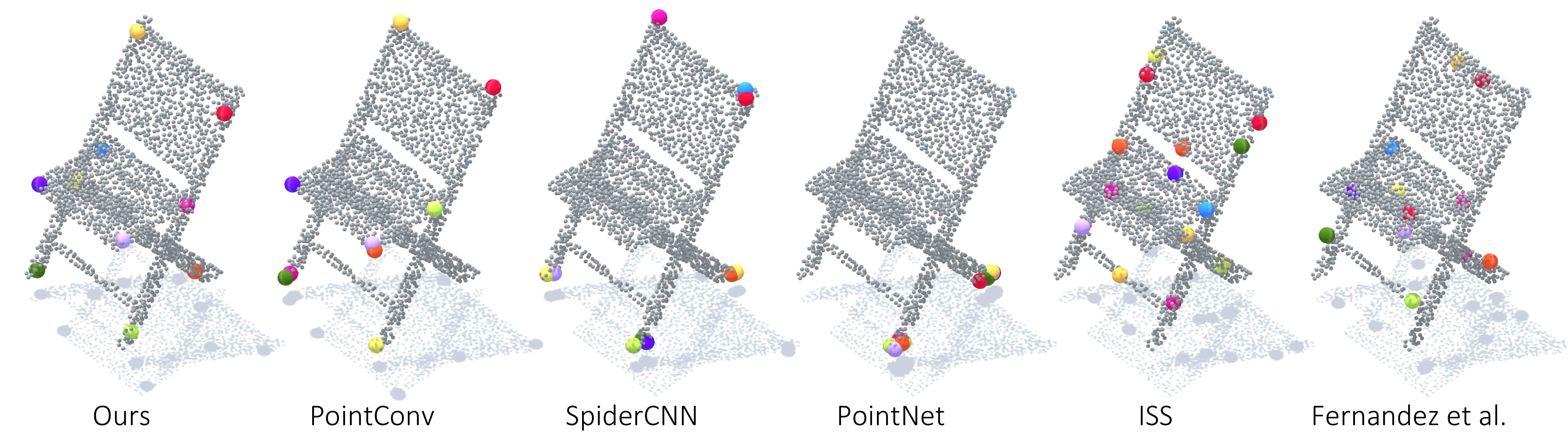}
  \caption{\textbf{Comparison between different supervised and unsupervised detectors.} Our detector produces salient, semantically rich keypoints and is comparable to or even outperforms supervised methods, as shown.}
  \label{salcomp}
\end{figure*}

\subsection{Evaluation on human annotated data}
We first evaluate saliency and alignment of keypoints detected by the proposed Skeleton Merger. More specifically, we consider evaluation on two large-scale keypoint datasets, KeypointNet\cite{you2020keypointnet} and data from SyncSpecCNN \cite{yi2016scalable}, where keypoints are annotated by experts with semantic correspondence labels.

\paragraph{Keypoint saliency}
For keypoint saliency, we compute the mean Intersection over Unions (mIoU) \cite{teran20143d} metric, where
\begin{equation}
\mathit{IoU} = \frac{\mathit{TP}}{\mathit{TP} + \mathit{FP} + \mathit{FN}}.
\end{equation}

mIoU is computed with a threshold of 0.1 in euclidean distance. Skeleton Merger is trained and tested on KeypointNet. Also, several supervised networks, PointNet \cite{qi2017pointnet}, SpiderCNN \cite{xu2018spidercnn} and PointConv \cite{wu2019pointconv}, are trained on KeypointNet to predict the likeliness of each point to be a keypoint as a comparison. In addition, the recent unsupervised detector \cite{fernandez2020unsupervised} by Fernandez \etal \footnote{The method in \cite{fernandez2020unsupervised} requires a category-specific symmetry prior, which is not available in the Guitar class.} and two traditional methods, Harris3D \cite{sipiran2011harris} and ISS \cite{Zhong2009IntrinsicSS}, are also compared.  

The results are shown in Tab.\ \ref{tableMIOU}. It can be seen that in these categories where keypoints are (at least partially) well defined, our unsupervised detector shows competitive performance to, or even outperforms, the supervised networks, in terms of mIoU, and is by far superior to traditional methods in detecting salient keypoints.
A visualized comparison between different methods is shown in Fig.\ \ref{salcomp}.

\begin{table}[h]
  \begin{center}
  \begin{tabular}{llll}
    \toprule
    \     & Airplanes & Chairs & Guitars \\
    \midrule
    PointNet & 45.4 & 23.8 & 0.2 \\
    SpiderCNN & 55.0 & 49.0 & 17.0 \\
    PointConv & 93.5 & 86.0 & 84.9 \\
    \midrule
    Harris3D & 42.8 & 15.1 & 33.1 \\
    ISS & 36.3 & 11.6 & 37.0 \\
    \midrule
    Fernandez \etal & 69.7 & 51.2 & - \\
    \midrule
    Ours & 79.4 & 68.4 & 55.0 \\
    \bottomrule
  \end{tabular}
  \end{center}
  \caption{mIoU scores of Skeleton Merger, different supervised networks, method of Fernandez \etal and traditional keypoint detectors on KeypointNet.}
  \label{tableMIOU}
\end{table}

\paragraph{Keypoint alignment}
For keypoint alignment, Skeleton Merger is trained on the ShapeNetCoreV2 \cite{shapenet2015} dataset, and DAS described in the previous section is evaluated on the KeypointNet\cite{you2020keypointnet} and SyncSpecCNN \cite{yi2016scalable} datasets. The results are shown in Tab.\ \ref{dast}. We compare the DAS scores with the method of Fernandez \etal \cite{fernandez2020unsupervised}.

\begin{table}[h]
  \begin{center}
  \begin{tabular}{llll}
    \toprule
    \     & Airplane (K) & Chair (K) & Chair (S) \\
    \midrule
    Fernandez \etal & 61.4 & 64.3 & 54.2 \\
    Ours & \textbf{77.7} & \textbf{76.8} & \textbf{73.8} \\
    \bottomrule
  \end{tabular}
  \end{center}
  \caption{DAS scores of Skeleton Merger and method of Fernandez \etal on KeypointNet (K) and SyncSpecCNN (S).}
  \label{dast}
\end{table}

\begin{table*}
  \begin{center}
  \begin{tabular}{l|ll|ll}
    \hline
    \ & \multicolumn{2}{c|}{mIoU} & \multicolumn{2}{c}{DAS} \\
    \ & Airplane & Chair & Airplane & Chair \\
    \hline
    Full Skeleton Merger & 79.4 & \textbf{68.4} & \textbf{77.7} & \textbf{76.8} \\
    No activation strengths & 55.5 & 8.4 & 72.1 & 65.2 \\
    No fidelity loss & 78.2 & 60.0 & 76.2 & 74.4 \\
    No coverage loss & 17.8 & 1.1 & 35.6 & 37.3 \\
    No offsets & \textbf{85.6} & 62.0 & 72.4 & 75.6 \\
    \hline
  \end{tabular}
  \end{center}
  \caption{Ablation study of different components in Skeleton Merger.}
  \label{ablt}
\end{table*}

\begin{figure*}[ht]
  \includegraphics[width=0.8\linewidth]{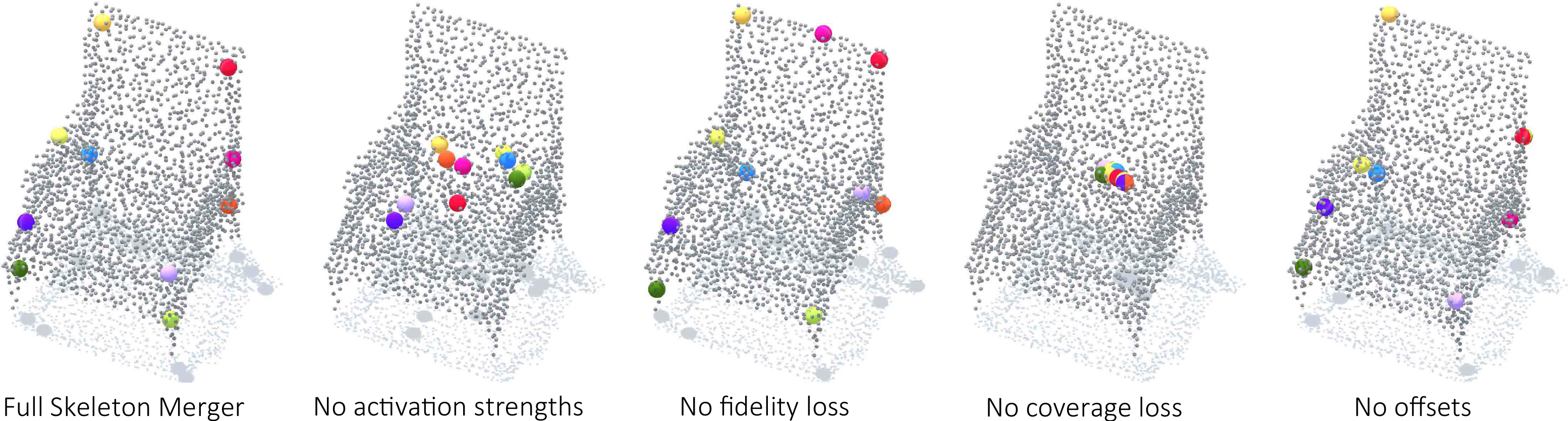}
  \centering
  \caption{\textbf{Visualization results of the ablation study.} Minor degeneracies can be seen in models without offsets or the fidelity loss. Major performance drop is seen if either the activation strengths or the coverage loss is removed.}
  \label{ablvis}
\end{figure*}

\subsection{Repeatability}
In this section we investigate the repeatability of detected keypoints to Gaussian additive noise and point cloud downsampling. Gaussian noises of different strengths and different sampling ratios are applied on the point clouds, and the same Skeleton Merger network trained on ShapeNetCoreV2 \cite{shapenet2015} is applied for keypoint detection.

The keypoints that are detected on these modified point clouds are compared with those detected from the clean, original point cloud \emph{in order}, that is, keypoints are compared one-by-one as a list instead of a set as in most previous works. If the distance between a keypoint detected from the original point cloud and a keypoint detected under noise or subsampling is smaller than $10\%$ of the model size, the keypoint is considered repeatable.

\begin{figure}[ht]
  \includegraphics[width=0.49\linewidth]{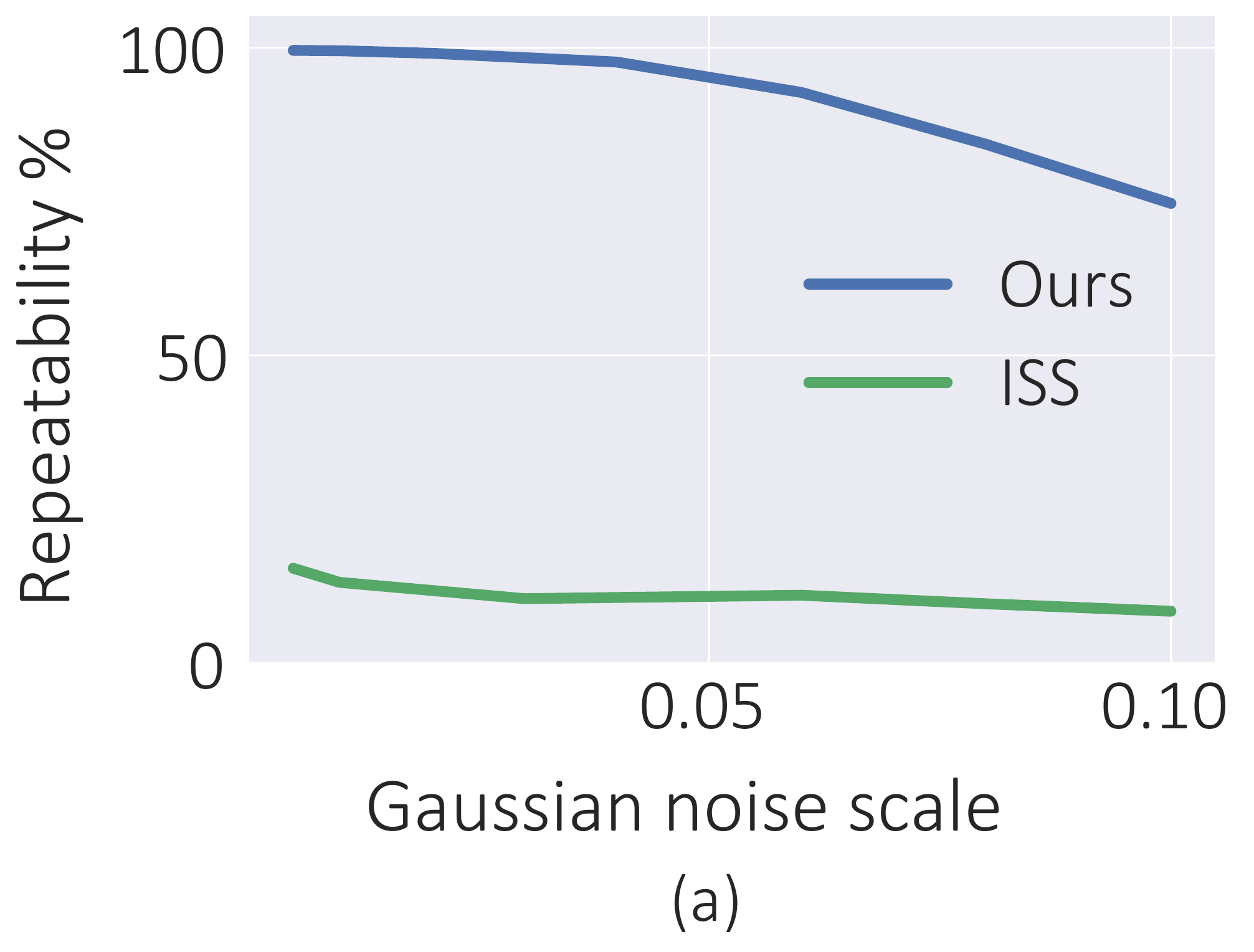}
  \includegraphics[width=0.49\linewidth]{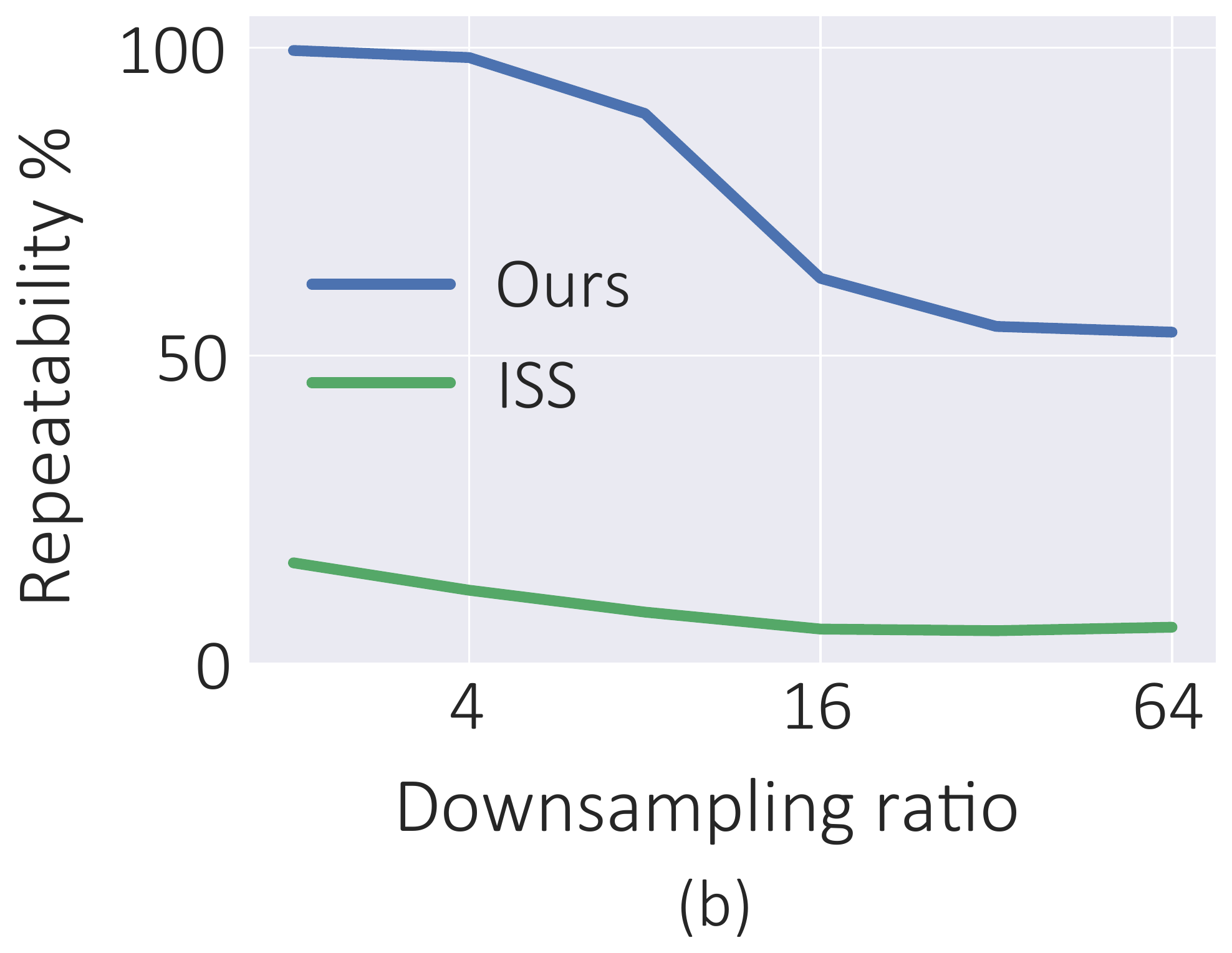}
  \includegraphics[width=0.93\linewidth]{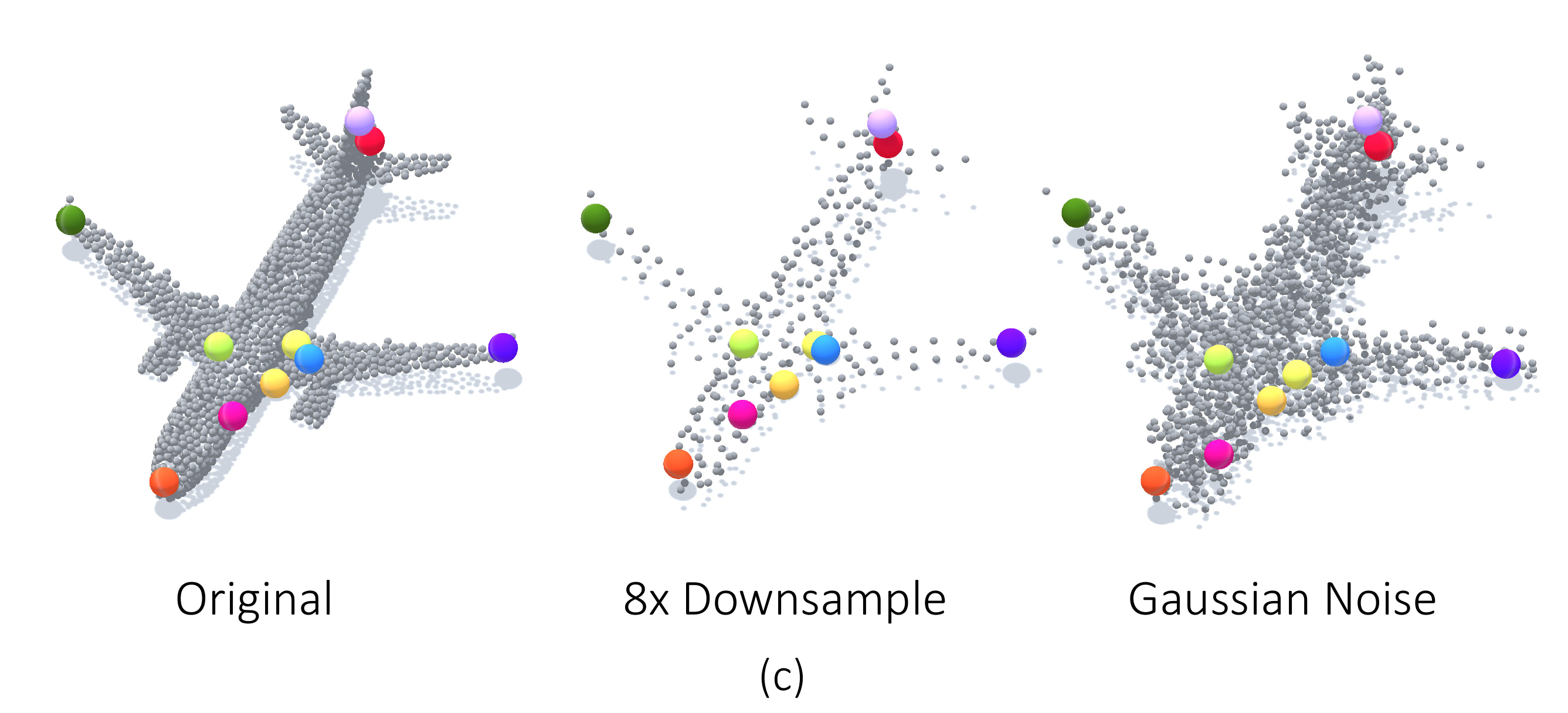}
  \centering
  \caption{\textbf{Repeatability of Skeleton Merger and ISS on ShapeNetCoreV2.} (a) Repeatability under Gaussian noise. (b) Repeatability under downsampling. (c) Visualization results of keypoints from Skeleton Merger under the scenarios. Downsample rate is 8x and the Gaussian noise scale is $0.05$.}
  \label{repfig}
\end{figure}

The results are shown in Fig.\ \ref{repfig}. We compare the results with ISS \cite{Zhong2009IntrinsicSS}. It can be seen that the aligned keypoints detected remain highly repeatable under Gaussian noise or downsampling. They also stay well-aligned.

\subsection{Ablation study}
We carried out experiments on the effectiveness of proposed components in Skeleton Merger. Tab.\ \ref{ablt} and Fig.\ \ref{ablvis} shows the results from the network with different settings. The models are trained on ShapeNetCoreV2 \cite{shapenet2015} and evaluated on KeypointNet \cite{you2020keypointnet}.

\paragraph{Activation strengths} Without activation strengths, \ie all activations of sub-clouds are set to 1, the network has to cope with the non-existing edges to fit the input point clouds, and there lacks a mechanism to enforce alignment of keypoints (the skeleton topology is no longer utilized for alignment), thus the full model outperforms model of this version both in saliency (mIoU) and alignment (DAS).

\paragraph{Trainable offsets} Without offsets, only straight lines between keypoints are allowed. These lines fit the shape of airplanes well, and removing offsets reduces variance in the model, so this version outperforms the full model in terms of keypoint saliency. In other cases, however, the model without offsets is not so lucky. It suffers from filling squares, balls and other shapes with straight lines.

\paragraph{Composite Chamfer Distance} The Composite Chamfer Distance is at the core of the training process.

Without the fidelity loss, activation strengths soon go to 1 because of the stop-iteration procedure in the coverage loss. As a result, the network can only learn meaningful latents in the first several epochs, and stops improving due to the same reasons without activation strengths, causing degeneracies in performance.

Without the coverage loss, activation strengths soon go to 0 as fidelity loss is minimized, preventing the network from learning anything meaningful. This also emphasizes the importance of coverage in keypoint detection.

\subsection{Qualitative results and limitations}

\begin{figure}[ht]
  \includegraphics[width=\linewidth]{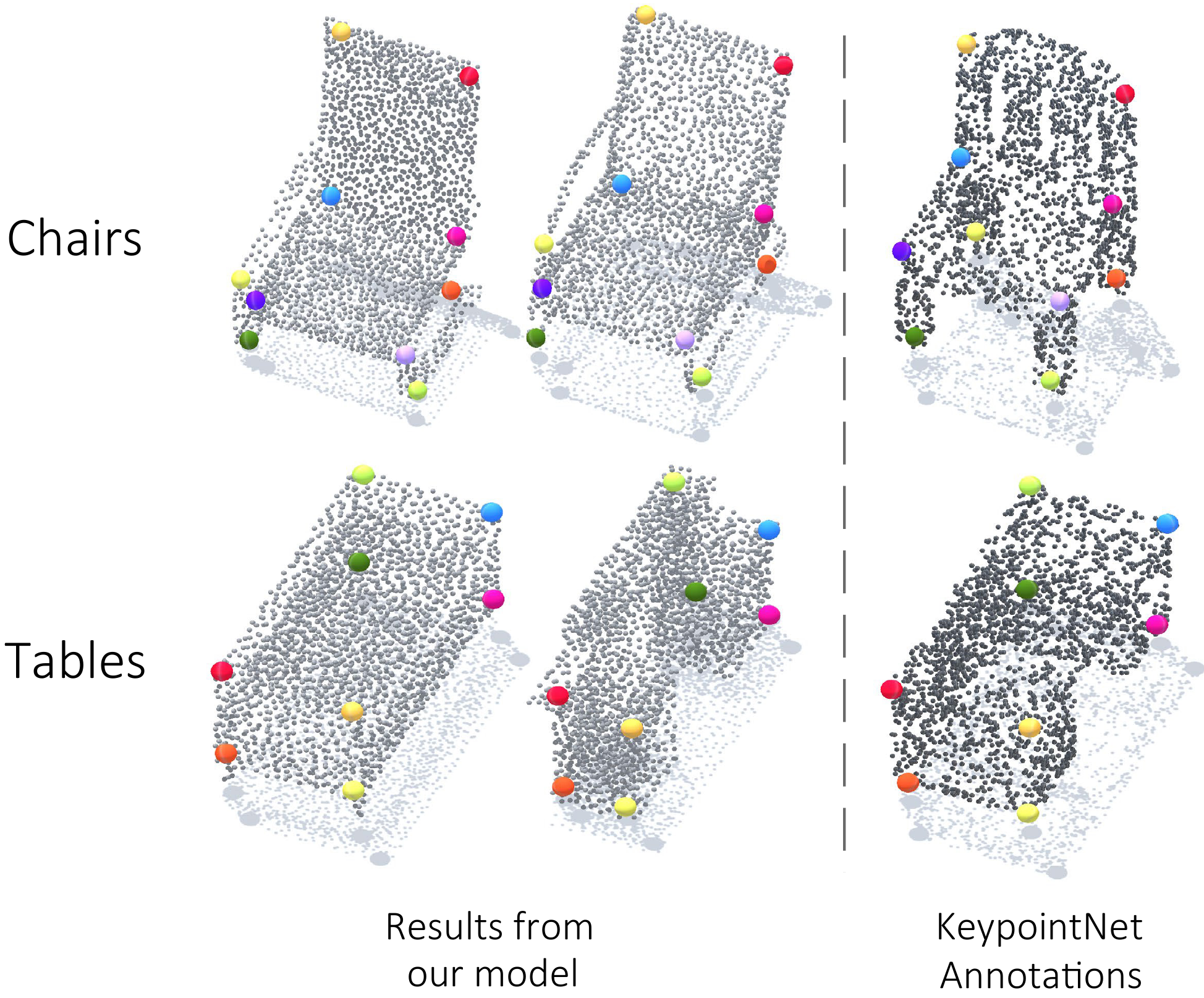}
  \centering
  \caption{\textbf{Examples of keypoints detected by Skeleton Merger on ShapeNetCoreV2.} Keypoints detected on chairs and tables, together with a set of annotations from KeypointNet are displayed.}
  \label{fvis_ct}
\end{figure}

\begin{figure}[ht]
  \includegraphics[width=\linewidth]{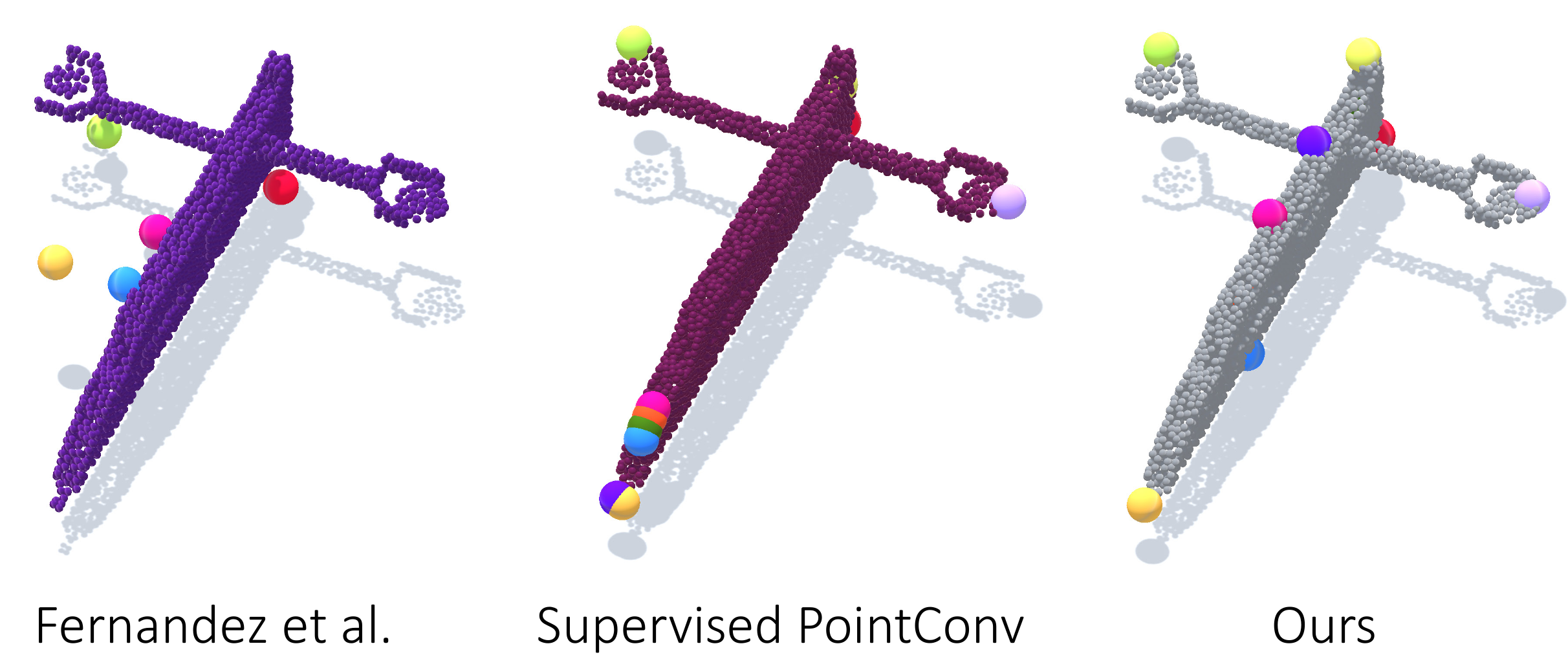}
  \centering
  \caption{\textbf{Comparison of different detectors on an irregular-shaped airplane.} The method from Fernandez \etal fails to generalize on this instance, and supervised PointConv shows degeneracies, while our method works well.}
  \label{irrplane}
\end{figure}

In this section, we give some qualitative results of our method and discuss the keypoint ambiguity and limitations.

\begin{figure}[htb]
  \includegraphics[width=\linewidth]{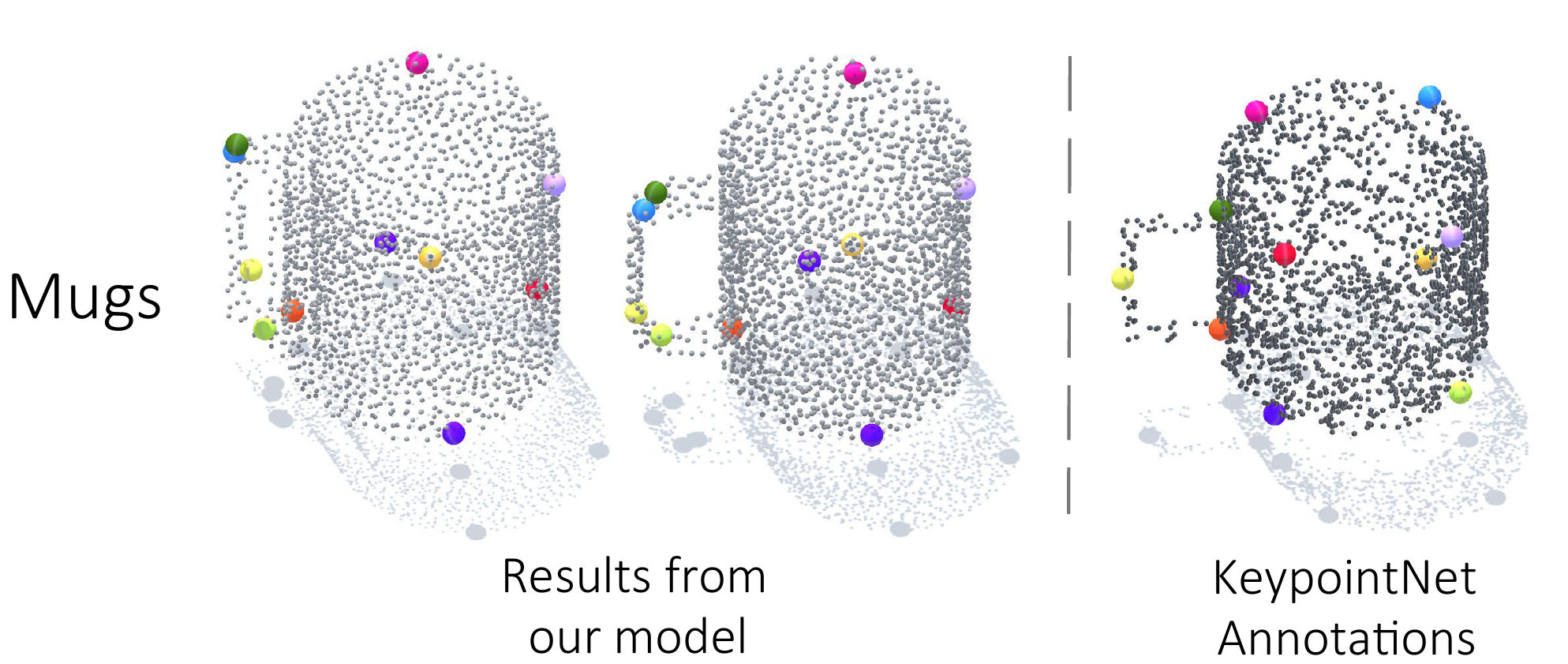}
  \centering
  \caption{\textbf{Examples of keypoint detections and human annotations in the mug category.} The points on the lip of the mugs share the same semantics, which leads to ambiguity in keypoint definition.}
  \label{fvis_am}
\end{figure}

\paragraph{Qualitative visualization results} Fig.\ \ref{fvis_ct} shows some visualization results of keypoints detected by Skeleton Merger on different object categories. It can be seen that these points are well-aligned between different instances of objects and cover most points with semantics in the point cloud. They correspond well to the human annotations.

\paragraph{Irregular-shaped instances} It is worth mentioning the generalization capability of our keypoint detector to irregular-shaped objects in an category. As shown in Fig.\ \ref{irrplane}, the model of Fernandez \etal \cite{fernandez2020unsupervised} fails to generalize to the irregular-shaped airplane, and the supervised PointConv \cite{wu2019pointconv} network shows some minor degeneracies in keypoint detection due to low frequencies of irregular objects appearing in the training set. Our method still works fine on this irregular-shaped instance.

\paragraph{Keypoint ambiguity} It is demonstrated in Fig.\ \ref{fvis_am} that keypoint definitions are ambiguous in some objects. Points on the lip of the mug, for example, are equivalent in semantics due to circular symmetry of the shape. Skeleton Merger and the KeypointNet ground truth both yield symmetric points, but points with different angles are selected.

The ambiguity makes it hard to aggregate human annotations to obtain a high-quality dataset for a wide range of objects, such as jars and cameras. As discussed before, this imposes a strong limit on the application of supervised methods for keypoint detection.

\paragraph{Limitations} Skeleton Merger is capable of generating semantically-rich and well-aligned keypoints. However, it is less sensitive to local semantics than global coverage. For example, joints are already covered by a cross of two skeleton edges. Selecting keypoints at these points may not reduce the global losses.

\section{Conclusion}
In this paper, we present Skeleton Merger, an unsupervised aligned keypoint detector. Composite Chamfer Distance (CCD) is proposed as a loss function to guide the network to detect high-quality keypoints by reconstructing a point cloud through refining its skeleton. Evaluations are performed on the quality and repeatability of detected keypoints. Our detector shows impressive performance detecting salient and well-aligned keypoints.

\section*{Acknowledgement}
This work is supported in part by the National Key R\&D Program of China, No.\ 2017YFA0700800, National Natural Science Foundation of China under Grants 61772332, SHEITC (2018-RGZN-02046) and Shanghai Qi Zhi Institute.

\newpage
{\small
\bibliographystyle{ieee_fullname}
\bibliography{egbib}
}
\end{document}